\documentclass[conference]{IEEEtran}
\IEEEoverridecommandlockouts

\usepackage{cite}
\usepackage{amsmath,amssymb,amsfonts}
\usepackage{algorithmic}
\usepackage{graphicx}
\usepackage{textcomp}
\usepackage{xcolor}

\usepackage{lipsum}
\usepackage{lettrine}
\usepackage{hyperref}
\usepackage{soul,color}
\usepackage[linesnumbered,ruled,vlined]{algorithm2e}
\usepackage{booktabs}
\usepackage{multirow}
\usepackage{graphicx}
\usepackage{subcaption}
\usepackage{caption}

\def\BibTeX{{\rm B\kern-.05em{\sc i\kern-.025em b}\kern-.08em
    T\kern-.1667em\lower.7ex\hbox{E}\kern-.125emX}}

\begin{document}

\title{Modeling Romanized Hindi and Bengali: Dataset Creation and Multilingual LLM Integration

% IndoTranslit: A Large-Scale Transliteration Dataset and Multilingual LLM for Hindi and Bengali

% A Transliteration Dataset and Multilingual LLM for Romanized Hindi and Bengali Script

\thanks{
\textsuperscript{*}Both author contributed equally to this work. \\
Full code and dataset available at: \url{https://github.com/sk-research-community/multilingual-transliterator-llm-training}
}
}

%  ------------------------------ 
%  Other Title Options...

% Roman2Indic: A Multilingual Dataset and Marian-Based Transliteration Model for Hindi and Bengali Romanized Scripts

% IndoScript: Transliteration Dataset and Multilingual Language Model for Hindi and Bengali Romanized Texts
%  ------------------------------ 

\author{
Kanchon Gharami\textsuperscript{1*}, 
Quazi Sarwar Muhtaseem\textsuperscript{2*}, 
Deepti Gupta\textsuperscript{3}, 
Lavanya Elluri\textsuperscript{3}, 
Shafika Showkat Moni\textsuperscript{1} \\[0.8em]
\textsuperscript{1}\textit{Department of EECS, Embry-Riddle Aeronautical University, Daytona Beach, FL, USA} \\
\textsuperscript{2}\textit{NLP Engineering Team, Hishab Singapore Pte. Ltd, Singapore} \\
\textsuperscript{3}\textit{Department of Computer Information Systems, Texas A\&M University - Central Texas, Texas, USA} \\
Emails: gharamik@my.erau.edu, sarwarshafee@gmail.com, d.gupta@tamuct.edu, elluri@tamuct.edu, monis@erau.edu
}

% \author{
% \IEEEauthorblockN{Kanchon Gharami\textsuperscript{*}}
% \IEEEauthorblockA{\textit{Department of Computer Science and Engineering} \\
% \textit{Embry-Riddle Aeronautical University}\\
% Daytona Beach, FL, USA \\
% gharamik@my.erau.edu}
% \and
% \IEEEauthorblockN{Quazi Sarwar Muhtaseem\textsuperscript{*}}
% \IEEEauthorblockA{\textit{NLP Engineering Team} \\
% \textit{Hishab Singapore Pte. Ltd}\\
% Singapore \\
% sarwarshafee@gmail.com}
% \and
% \IEEEauthorblockN{Deepti Gupta}
% \IEEEauthorblockA{\textit{Department of Computer Information Systems} \\
% \textit{Texas A\&M University - Central Texas}\\
% Texas, USA\\
% d.gupta@tamuct.edu}
% \and

% \IEEEauthorblockN{Lavanya Elluri}
% \IEEEauthorblockA{\textit{Department of Computer Information Systems} \\
% \textit{Texas A\&M University - Central Texas}\\
% Texas, USA\\
% elluri@tamuct.edu}
% \and

% \IEEEauthorblockN{Shafika Showkat Moni}
% \IEEEauthorblockA{\textit{Department of Computer Science and Engineering} \\
% \textit{Embry-Riddle Aeronautical University}\\
% Daytona Beach, FL, USA \\
% monis@erau.edu}
% }

\maketitle

\begin{abstract}
The development of robust transliteration techniques to enhance the effectiveness of transforming Romanized scripts into native scripts is crucial for Natural Language Processing tasks, including sentiment analysis, speech recognition, information retrieval, and intelligent personal assistants. Despite significant advancements, state-of-the-art multilingual models still face challenges in handling Romanized script, where the Roman alphabet is adopted to represent the phonetic structure of diverse languages. Within the South Asian context, where the use of Romanized script for Indo-Aryan languages is widespread across social media and digital communication platforms, such usage continues to pose significant challenges for cutting-edge multilingual models. While a limited number of transliteration datasets and models are available for Indo-Aryan languages, they generally lack sufficient diversity in pronunciation and spelling variations, adequate code-mixed data for large language model (LLM) training, and low-resource adaptation. To address this research gap, we introduce a novel transliteration dataset for two popular Indo-Aryan languages, Hindi and Bengali, which are ranked as the 3rd and 7th most spoken languages worldwide. Our dataset comprises nearly 1.8 million Hindi and 1 million Bengali transliteration pairs. In addition to that, we pre-train a custom multilingual seq2seq LLM based on Marian architecture using the developed dataset. Experimental results demonstrate significant improvements compared to existing relevant models in terms of BLEU and CER metrics.

\end{abstract}


\section{Introduction}
\label{sec:introduction}
With the advancement of large language models (LLMs), there has been an unprecedented expansion in natural language applications, ranging from virtual assistants and content generation to sentiment analysis and question answering. As LLMs become the conversational layer of the modern internet, their ability to interpret, retrieve, and reuse information from diverse linguistic inputs is becoming increasingly critical. Even the most advanced multilingual large language models face challenges in handling Romanized scripts, wherein non-English languages are represented using the 26 characters of the Roman alphabet.

% Still the best multilingual LLMs struggle with Romanized script—words from non-English languages written using the 26 letters of the Roman alphabet. 

\begin{figure}[ht]
    \centering
    \includegraphics[width=\linewidth]{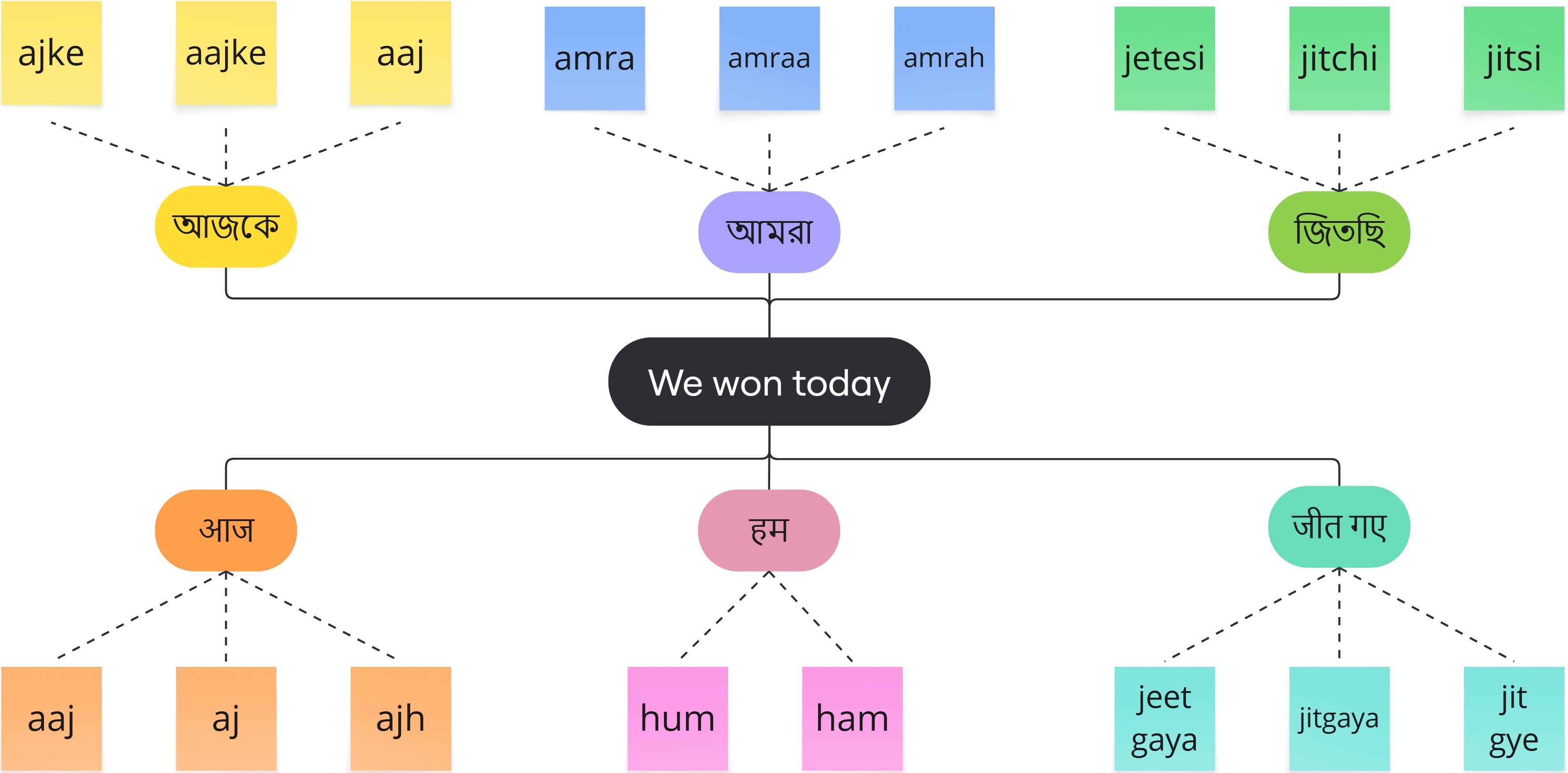}
    \caption{Variations in Romanizing words for same sentence}
    \label{fig:Transliteration_Example}
\end{figure}

The main reasons are the lack of grammar rules and the absence of structured training data. There is no large, well-organised corpus that captures the real spelling patterns people use, so the models have little to learn from. The problem is especially clear in South Asia, where Hindi and Bengali—spoken by about 609 million and 284 million people respectively, or roughly 11\% of the world’s population—are often typed in Roman script~\cite{wikilangspeakers, ammon2010world}. A 2014 study analyzing Hindi-English code-mixed social media content found that 93.17\% of native Hindi Speaker's posts were written in Roman script, while only 2.93\% used Devanagari~\cite{vyas2014pos}. Additionally, a 2024 study of Indian Twitter users revealed that the proportion of users preferring Hinglish (code-mixed Hindi-English) increased from 44.9\% in 2014 to 56.3\% after 2020, with a steady annual growth rate of 1.2\%~\cite{sengupta2024social}. As a result, billions of words reach LLM back-ends each day in a form they were never trained to understand. This leads to poor performance in tasks like search, sentiment analysis, and conversational responses for millions of users.

The preference for Roman script is partly ergonomic. Devanāgarī (Hindi alphabets) encodes 47 base letters plus vowel marks and conjuncts, while the Bengali alphabet adds 50 letters with their richly enriched set of diacritics. Mapping these richly shaped, context-sensitive glyphs onto a 26-key QWERTY layout demands complex input methods and frequent mode switches, ultimately making the writing experience very slow. This limitation provokes the region’s mobile-first internet users to adopt an informal orthography, writing native words using the Roman alphabet based on pronunciation. Over time, this has evolved into a grammarless and highly flexible form of Romanized text, commonly known as Banglish for Bengali and Hindish for Hindi.. Since it tries to represent a large number of native glyphs using only 26 Roman symbols, the result is a loosely structured script that appears almost like hieroglyphics in modern multilingual LLMs which is difficult to interpret, normalize, or learn from.

Automatic transliteration~\cite{karimi2011machine, mammadzada2023review, deselaers2009deep}, which transforms Romanized text back into native script, can mitigate this existing research gap. Figure~\ref{fig:Transliteration_Example} provides more examples that reflect real-world variation, where a single native phrase may appear in multiple Romanized forms depending on the user's spelling habits, phonetic interpretation, or typing speed. This inconsistency makes transliteration a non-trivial and essential pre-processing step for enabling downstream tasks.

% \subsection{Our Contributions}
% \label{subsec:our_contributions}
To address these issues, our work introduces a scalable, multilingual solution for transliterating Romanized Indo-Aryan script. The key contributions are as follows:

\begin{itemize}
    \item IndoTranslit Dataset: We develop a novel large-scale transliteration dataset with over \textbf{2.7 million pairs} (\textbf{1.8M Hindi} and \textbf{975k Bengali}) making it  the largest benchmarks for Romanized Indo-Aryan script. The dataset includes both clean phonetic forms and  noisy user-generated variants.
    
    \item Multilingual Model Training: We \textbf{pre-train} a 60M-parameter Marian-based Seq2Seq LLM capable of handling both Hindi and Bengali languages with a shared tokenizer.

    \item Low-Resource Adaptation: Our model is optimized for deployment in \textbf{low-resource environments}, with limited memory and fast training/inference on standard GPU servers.

    \item Performance Gains: Our approach outperforms strong baselines on BLEU and CER metrics, especially for \textbf{noisy and code-mixed (multilingual)} input.

    \item Robustness Across Lengths: We analyze sentence-length vs BLEU trends and show consistent performance across short and long inputs.

    \item Open Resources: We \textbf{open-source the dataset}, model, and training scripts to support reproducibility and downstream use.
\end{itemize}

The rest of the paper is organized as follows. Section~\ref{sec:related_work} present existing research on transliteration with their pros and cons. Section~\ref{sec:dataset} describes the dataset collection and processing methodology. Section~\ref{sec:model} details the model architecture and training setup. Section~\ref{sec:results} presents evaluation results and analysis. Finally, we conclude the paper in Section~\ref{sec:conclusion}.

\section{Related Work}
\label{sec:related_work}
In this sub-section, we review benchmark datasets and relevant approaches, highlighting their contributions, strengths, and limitations.

Fahim et al. introduced BanglaTLit~\cite{fahim2024banglatlit}, a large-scale transliteration dataset for Bengali containing 42.7k annotated pairs and a 245k-sample Romanized corpus for pretraining. They manually curated the dataset from diverse sources such as TrickBD, Facebook, and YouTube, and further validated it with expert reviewers. The authors also proposed dual encoder-decoder models, combining pretrained encoders with BanglaT5~\cite{bhattacharjee2023banglanlg}, achieving strong results on back-transliteration benchmarks. Although the model performs well and contains data diversity, the amount of data is really small, the focus is limited to Bengali only, and coverage of the broader Indo-Aryan language diversity is missing.

Madhani et al. released Aksharantar~\cite{madhani2022aksharantar}, the largest public Indic transliteration dataset to date, with 26 million word-level pairs spanning 21 languages and 12 scripts. Data was mined from Wikidata, Samanantar, and IndicCorp, supplemented by 100k manually annotated samples. The authors also proposed IndicXlit, a multilingual transformer-based transliteration model, which achieved a 15\% accuracy gain over prior baselines on the Dakshina~\cite{roark2020processing} test set. The strength of this work lies in its multilingual scope and open-source resources, but it focuses mainly on canonical word-level transliteration rather than noisy, real-world inputs.

Andreas et al.~\cite{chari2025lost} studied how transliteration affects neural information retrieval. They found that strong models like BGE-M3~\cite{chen2024bge} perform poorly on Romanized queries. To solve this, they used a translate-train method that combines native and Romanized queries during training, which led to much better retrieval results. Haider et al.~\cite{haider2024banth} introduced BanTH, a 37.3k-sample hate-speech dataset in Romanized Bengali with multi-class labels. It shows that TB-Encoders pretrained on transliterated text enhance performance on noisy inputs, highlighting the value of transliteration-aware models.

The IndoNLP 2025~\cite{sumanathilaka2025indonlp} introduced benchmark tasks for Indic NLP, including transliteration, translation, and code-mixing. It provided unified datasets and evaluation tracks using multilingual models such as BERT and LLaMA, enabling consistent comparison across Indo-Aryan and Dravidian languages. The transliteration track offered both word- and sentence-level settings, promoting reproducibility and shared baselines. While the framework strengthened evaluation consistency, it relied on existing corpora like Dakshina and Aksharantar, limiting real-world coverage. Future work could expand toward informal and noisy text domains.

IndiDataMiner~\cite{kumar2025team} present a Hindi back-transliteration system built on the LLaMA 3.1 (8B) model using LoRA fine-tuning. The system converted Romanized Hindi to Devanagari at the sentence level, trained on the Dakshina dataset in an instruction-tuned setup. It achieved BLEU 0.8866 (character) and 0.6288 (word), outperforming IndicXlit baselines. The model proved robust to vowel omission but covered only structured Hindi, lacking social-media diversity. Future directions include multilingual expansion and lightweight deployment.

% The IndoNLP 2025~\cite{sumanathilaka2025indonlp} introduced several benchmark settings for Indic language tasks, including transliteration. One of their research teams, IndiDataMiner~\cite{kumar2025team}, focused on Roman-to-native transliteration using neural models. Their system utilized sub-word tokenization and shallow CNN encoders and reported strong performance in both constrained and unconstrained tracks. However, their dataset usage was limited to shared task data, and no new corpora were proposed.

\begin{table*}[ht]
\centering
\footnotesize
\caption{Existing Models Summary}
\label{tab:existing_model_summary}
\begin{tabular}{lllll}
\hline
\textbf{Paper} & \textbf{Model} & \textbf{Params} & \textbf{Focus} & \textbf{Cons} \\
\hline
Aksharantar & IndicXlit (Transformer) & 11M & Multilingual & Word-level only \\
BanglaTLit & TB+T5\_NMT & 110M & Bengali & Small-scale, single-lang \\
BANTH & TB-mBERT & 110M & Hate-speech, Bengali & Task-specific \\
IndoNLP & BERT, LLaMA & 8B & Multilingual & No new data, basic setup \\
IndiDataMiner & LLaMA 3.1 & 8B & Roman to Devanagari & Hindi only, no corpus \\

Assamese Social Media & BiLSTM, Transformer & 12M & Romanized text & Small, domain-limited \\
Context-Aware (Google) & LSTM, Transformer Ensembles & 500M+ & Sentence-level context & Heavy compute \\
Sinhala TU-Model & Encoder–Decoder (LSTM) & 15M & Contextual back-translit & Low domain generalization \\
GRT & Rule-based (Mapping) & -- & Grapheme-level & Noisy-text failure \\
Romanagari & fMRI Study (Cognitive) & -- & Reading load analysis & Small, lab-only \\
\hline
\end{tabular}
\end{table*}

\begin{table}[ht]
\centering
\caption{Benchmark Datasets Summary}
\label{tab:existing_dataset_summary}
% \scriptsize
\begin{tabular}{lp{1.6cm}ll}
\hline
\textbf{Paper} & \textbf{Languages} & \textbf{Size} & \textbf{Sources} \\
\hline
Aksharantar & 21 Indic langs & 26M & Wiki, IndicCorp, Manual \\
BanglaTLit & Bengali only & 280k & TrickBD, FB, Blogs, Wiki \\
BANTH & Bengali only & 37k & YouTube comments \\
IndoNLP & 5 Indic langs & 15k & Dakshina, Aksharantar \\
IndiDataMiner & Hindi only & 20k & Dakshina \\
% Dakshina~\cite{roark2020processing} & 12 Indic langs & 2M & Wiki, Native–Roman pairs \\
Sinhala Romanized & Sinhala only & 35k & Social media, forums \\
GRT & Punjabi & 65k & Local corpus \\
FIRE Shared Task & 5 Indic langs & 30k & FIRE dataset \\
\hline
\end{tabular}
\end{table}

Baruah et al.~\cite{baruah2024assamesebacktranslit} studied how Assamese users transliterate their language into Roman script on social media, a challenge absent in earlier work. They collected Romanized Assamese text from Facebook, YouTube, and X, manually aligned it with native script, and analyzed variation against six standard schemes. Three transliteration models: PBSMT, BiLSTM with attention, and Transformer were trained, where the BiLSTM model showed the best accuracy. The work offers the first systematic study for Assamese but is limited by dataset size and social-media bias, leaving room for larger multilingual expansion.

Kirov et al.~\cite{kirov2024context} proposed a context-aware transliteration framework for South Asian languages, improving over word-level models that ignore sentence context. Using the Dakshina dataset of 12 Indic languages, they trained ensemble LSTM and Transformer models combined with language modeling. Their approach achieved a 3.3\% absolute improvement in WER and tripled decoding speed. The study shows strong contextual benefits but is computationally heavy and still data-dependent, suggesting lighter multilingual extensions in future.

Nanayakkara et al.~\cite{nanayakkara2022context} presented a back-transliteration model for Romanized Sinhala text using an encoder–decoder framework with Transliteration Units (TUs) instead of characters. The TU-based representation improved contextual accuracy, achieving BLEU 0.81 and METEOR 0.78 on their primary dataset. While effective, the model’s performance dropped on unseen data, highlighting the need for larger, diverse training sources and transfer learning for domain adaptation.

Singh and Sachan~\cite{singh2019grt} introduced GRT, a Gurmukhi-to-Roman transliteration system covering all word types rather than just named entities. The model uses handcrafted character mapping and linguistic rules to perform accurate script conversion, achieving 99.27\% accuracy on 65k Punjabi words. The rule-based design ensures precision and clarity but lacks robustness for noisy or informal text, motivating hybrid or neural extensions.

Rao et al.~\cite{rao2013cost} analyzed the cognitive cost of reading Romanized Hindi (Romanagari) using fMRI. They found that Romanized text activates stronger phonological and attention-related brain regions, such as the left inferior parietal lobule and mid-cingulum, compared to native Devanagari or English. The findings highlight that transliteration demands extra neural effort, though results are limited to small, lab-based samples and need broader validation in multilingual contexts.

Prabhakar and Pal~\cite{prabhakar2018machine} provided a detailed survey of transliteration and transliterated text retrieval, outlining deterministic, probabilistic, and hybrid methods. They discussed challenges such as spelling variation, code-mixing, and lack of standard evaluation for Indian languages. The paper offers a clear taxonomy and context for modern systems but lacks empirical comparison. Future work should integrate deep learning and shared benchmarks for consistency across languages.

Sequiera et al.~\cite{sequiera2014word} developed a simple system for word-level language identification and back-transliteration of mixed Romanized Indic–English text. Using trigram-based language identification and rule-plus-dictionary transliteration, they achieved 70–80\% accuracy across five language pairs in the FIRE shared task. The model is lightweight and interpretable but struggles with complex code-mixing, suggesting neural sequence models for better generalization.

A summary of existing models with limitations and datasets  are provided in  Table~\ref{tab:existing_model_summary} and Table~\ref{tab:existing_dataset_summary}, respectively. Exploring these studies reveals that most existing transliteration systems either focus on single languages or rely on clean, well-aligned corpora. Very few address the challenges of noisy, informal Romanized text that commonly appears in social media or user-generated content. Multilingual benchmarks such as Aksharantar and IndoNLP provide standardized evaluation, yet they still depend heavily on structured datasets like Dakshina, limiting real-world generalization. Moreover, most neural models remain word-level and lack contextual adaptation across scripts and domains. These gaps highlight the need for a unified, context-aware transliteration approach that can handle informal spelling variations, language mixing, and cross-domain noise, an issue this paper aims to address.

\section{Dataset}
\label{sec:dataset}

\subsection{Dataset Collection Methodology}
\label{subsec:dataset_collection}

\subsubsection{\textbf{Hindi}}
\label{subsubsec:hindi}
To construct our Hindi transliteration dataset, we first generated high-quality Romanized text by following the International Alphabet of Sanskrit Transliteration (IAST) principles~\cite{dalwadi2022uast}. IAST provides a consistent mapping from Devanāgarī script to Roman characters based on pronunciation, ensuring that each phoneme is clearly and accurately represented. We applied these rules to a large collection of native Hindi text, using the Hindi portion of the IIT Bombay English–Hindi parallel corpus~\cite{kunchukuttan-etal-2018-iit}, which contains sentence-aligned translation pairs. We extracted only the Hindi sentences and applied the IAST principle based open-source \texttt{indic-trans}~\footnote{https://github.com/libindic/indic-trans}~\cite{bhat2014iiit} library to convert them into Romanized form. This process produced 1.66 million clean and phonetically faithful Hindi–Romanized pairs, forming core of our dataset.

While IAST-based transliteration ensures correctness, it does not capture the informal variation seen in real-world user-generated text. To introduce diversity in spelling, accent, and writing style, we generated an additional 150k Romanized samples using \texttt{Gemini-2.0 Flash Lite}~\footnote{https://cloud.google.com/vertex-ai/generative-ai/docs/models/gemini/2-0-flash-lite}. We prompted the model to simulate how Hindi speakers typically write Romanized text in casual settings like messaging or social media. This introduced natural variation in vowel omission, consonant simplification, and non-standard spellings.

 We added 10k transliteration pairs from the Dakshina dataset~\cite{roark2020processing} to increase variations in scripts and include human-curated examples. These examples are manually annotated and represent frequently used Hindi words, serving as clean validation and test references. Together, this three-part strategy allowed us to build a large-scale dataset that is both linguistically accurate and representative of real-world usage.

\subsubsection{\textbf{Bengali}}
\label{subsubsec:bengali}
We develop a rule-based heuristic system that maps Bengali graphemes to Roman script based on phonetic similarity to generate Bengali transliteration dataset. The system segments input text into transliteration units, applies symbol-to-symbol mappings, and resolves consonant clusters and vowel diacritics using context-aware rules. We applied this method to the Bengali Wikipedia dataset, resulting in 975,215 Bengali Romanized sentence pairs.

Let $\Sigma_B$ denote the set of Bengali graphemes and $\Sigma_R$ the Roman target alphabet. The transliteration function $\mathcal{T}: \Sigma_B^* \rightarrow \Sigma_R^*$ maps an input word $w = [s_1, \dots, s_n]$ to its Romanized form by joining the mapped units:
\[
\mathcal{T}(w) = \text{join}\left(\left[\texttt{MapUnit}(u_i)\right]_{i=1}^{n}\right).
\]
We define four core mapping functions: independent vowels ($\mathcal{V}$), consonants ($\mathcal{C}$), vowel diacritics ($\mathcal{D}$), and nasal signs ($\mathcal{N}$). Table~\ref{tab:mapping_heuristics} summarizes the phoneme-level mapping rules used in our system.

\begin{table}[ht]
    \centering
    \caption{Bengali-to-Roman transliteration mapping.}
    \label{tab:mapping_heuristics}
    \begin{tabular}{c}
        \includegraphics[width=0.95\linewidth]{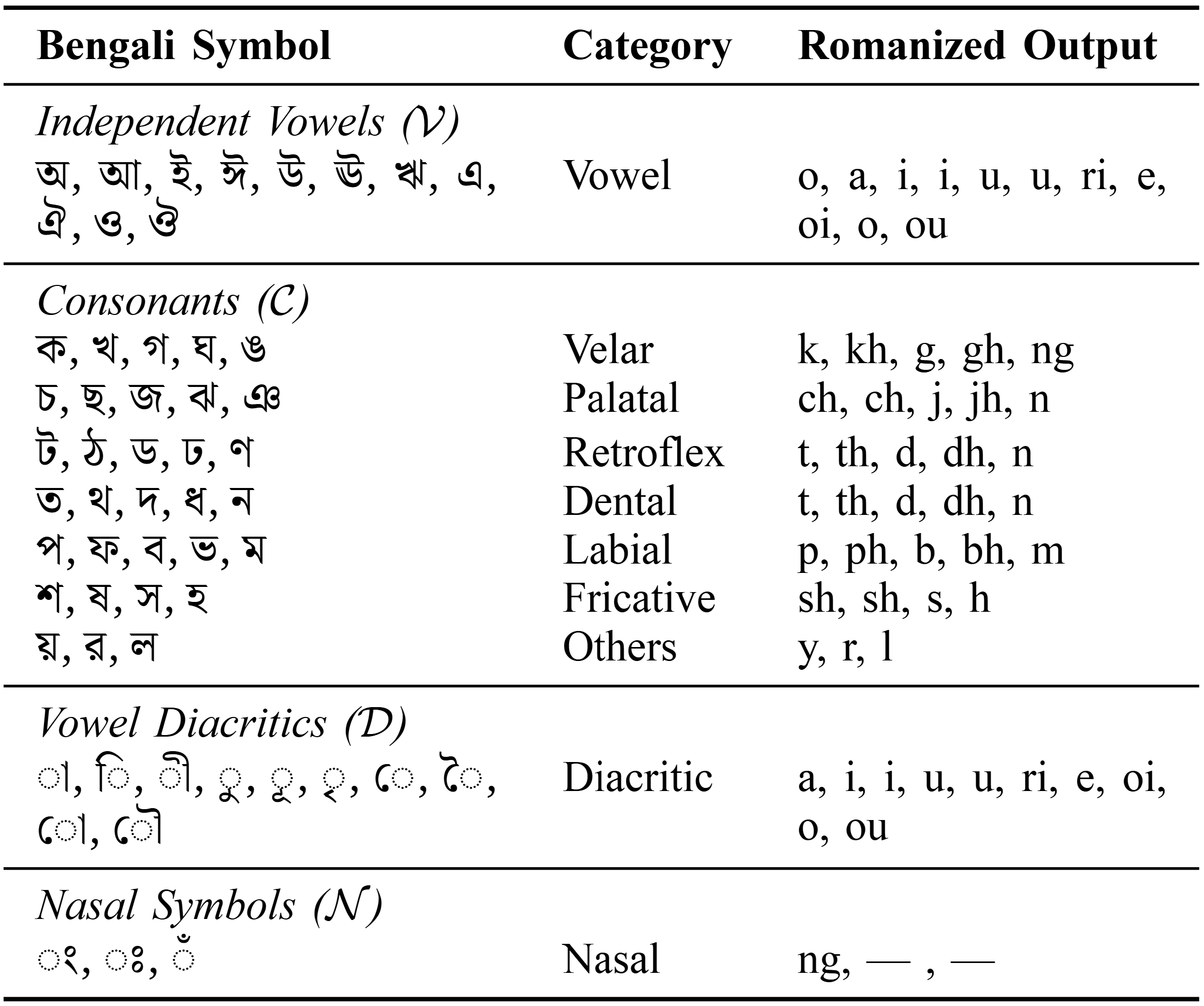}
    \end{tabular}
\end{table}

\begin{algorithm}[ht]
\footnotesize
\caption{Transliterate Bengali to Romanized}
\label{alg:translit}
\KwIn{Bengali word $w = [s_1, s_2, \dots, s_n]$}
\KwOut{Romanized string $r$}

$U \leftarrow$ \texttt{SplitUnits}($w$),\quad $r \leftarrow$ \texttt{""}\;

\For{$i \leftarrow 0$ \KwTo $|U|-1$}{
  $u \leftarrow U[i]$,\quad $t \leftarrow$ \texttt{MapUnit}($u$)\;

  \If{\texttt{IsConsonant}($u$) $\wedge$ $\neg$\texttt{HasDiacritic}($u$) $\wedge$ $\neg$\texttt{IsNasal}($u$)}{
    \If{$i = 0$ $\vee$ \texttt{IsConsonant}($U[i+1]$)}{
      $t \leftarrow t + \texttt{"a"}$\;
    }
  }

  $r \leftarrow r + t$\;
}
\Return{$r$}
\vspace{0.6em}

\SetKwFunction{FMap}{MapUnit}
\SetKwProg{Fn}{Function}{:}{}
\Fn{\FMap{$u$}}{
  \If{$u \in \mathcal{V}$}{\Return{\texttt{V[u]}}}
  \If{$u \in \mathcal{D}$}{\Return{\texttt{D[u]}}}
  \If{$u[0] \in \mathcal{C}$}{
    $t \leftarrow \texttt{C[u[0]]}$\;
    \For{$i \leftarrow 1$ \KwTo $|u|-1$}{
      \If{\texttt{u[i] == "virama"} $\wedge$ $u[i+1] \in \mathcal{C}$}{
        $t \leftarrow t + \texttt{C[u[i+1]]}$\;
      }
      \If{$u[i] \in \mathcal{D}$}{
        $t \leftarrow t + \texttt{D[u[i]]}$\;
      }
    }
    \Return{\texttt{t}}
  }
  \Return{\texttt{u}}
}
\end{algorithm}

\begin{table*}[ht]
\centering
\caption{Dataset statistics for IndoTranslit (Romanized Hindi and Bengali)}
\label{tab:dataset-stats}
\renewcommand{\arraystretch}{1.15}
\begin{tabular}{l|ccc|ccc|ccc|c}
\hline
\textbf{Language} & \multicolumn{3}{c|}{\textbf{Character Count}} & \multicolumn{3}{c|}{\textbf{Word Count}} & \multicolumn{3}{c|}{\textbf{Sentence Count}} & \textbf{Total Source Texts} \\
                 & \textbf{Mean} & \textbf{Max} & \textbf{Min} & \textbf{Mean} & \textbf{Max} & \textbf{Min} & \textbf{Mean} & \textbf{Max} & \textbf{Min} & \textit{(Before Cleaning)} \\
\hline
Hindi   & 70.70   & 8000   & 1     & 13.67   & 1380  & 1     & 1.06    & 218   & 1     & 1,178,149 \\
Bengali & 352.58  & 50655  & 50    & 50.63   & 7227  & 1     & 1.16    & 1110  & 1     & 957,948   \\
\hline
\textbf{Total Pairs} & \multicolumn{10}{c}{Hindi: \textbf{1,776,353} \hfill Bengali: \textbf{975,190}} \\
\hline
\end{tabular}
\end{table*}

\begin{table*}[ht]
\centering
\caption{Detailed Dataset Composition for IndoTranslit}
\label{tab:dataset_composition}
\footnotesize
\begin{tabular}{llllll}
\hline
\textbf{Language} & \textbf{Source} & \textbf{Pairing / Conversion Method} & \textbf{Rows} & \textbf{Cleaning Method} \\
\hline
Hindi & IITB / CFILT Corpus & Indic-trans Romanization & 1.66M & Rule-based normalization, script filter \\
Hindi & Gemini-2.0 Flash Model & Synthetic Romanization & 150k & Prompt-based generation, human verification  \\
Hindi & Dakshina Subset & Romanized benchmark pairs & 10k & Pre-aligned, token normalization \\
Hindi & Others open source & User-generated Romanized text & 15K & Manual curation, spelling normalization \\
Bengali & Wikipedia Paragraphs & Heuristic Romanization & 975k & Heuristic rule-based transliteration \\
\hline
\end{tabular}
\end{table*}

The transliteration process is described in Algorithm~\ref{alg:translit}. Given a Bengali word, we first segment it into transliteration units using a custom rule engine that detects vowels, consonants, virama-based conjuncts, and diacritics. Each unit is then mapped to its Roman equivalent using predefined heuristics. Additional rules append an implicit schwa (‘a’) when needed to match expected pronunciation patterns. 

This algorithm was applied paragraph-wise to filtered Wikipedia articles. Lines containing less than 50 characters or markup were discarded. Final transliteration pairs were saved in CSV format with optional case formatting for sentence-initial tokens.

\subsection{Dataset Characteristics and Statistics}
\label{subsec:dataset_statistics}

We perform a detailed analysis of the dataset to better understand its distribution and inform our modeling choices. Table~\ref{tab:dataset-stats} summarizes key statistics, including character and word counts, sentence-level aggregation, and vocabulary diversity for both the Hindi and Bengali data samples.

The dataset composition, detailed in Table~\ref{tab:dataset_composition}, outlines the balance between clean, synthetic, and user-generated Romanized text for both Hindi and Bengali. This mixture ensures coverage across formal and informal writing styles, allowing the model to generalize effectively from structured transliteration pairs to noisy real-world inputs.

\begin{figure}[ht]
    \centering
    \includegraphics[width=\linewidth]{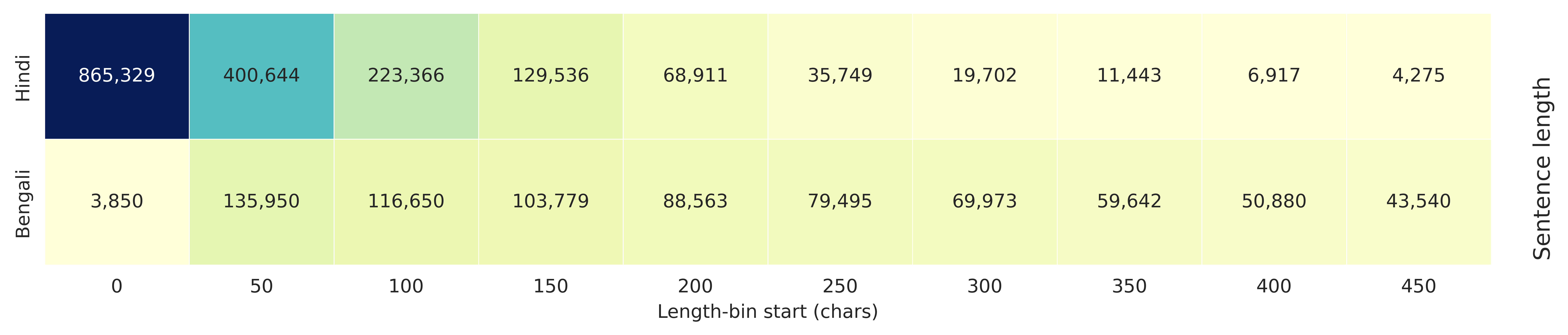}
    \caption{Distribution of character and word counts in Hindi and Bengali transliteration datasets.}
    \label{fig:len_heatmap}
\end{figure}

Figure~\ref{fig:len_heatmap} presents the distribution of sentence lengths across 50-character bins. Hindi sentences are generally shorter, with over 80\% concentrated below 150 characters. In contrast, Bengali sentences are notably longer and more varied, with a substantial portion exceeding 200 characters. This disparity suggests the need for models capable of handling longer input sequences and increased token density in Bengali.

\begin{figure}[ht]
    \centering
    \includegraphics[width=\linewidth]{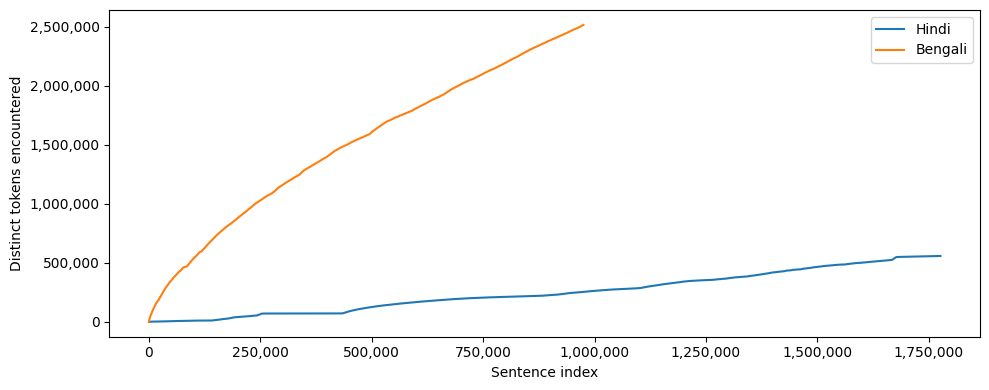}
    \caption{Vocabulary growth: cumulative number of distinct tokens observed over the dataset.}
    \label{fig:vocab_growth}
\end{figure}

To assess lexical diversity, we compute vocabulary growth curves by plotting the cumulative number of distinct tokens encountered with increasing sentence index, as shown in Figure~\ref{fig:vocab_growth}. Bengali exhibits significantly faster vocabulary expansion, crossing 2.5 million unique tokens after one million sentences, which is more than four times higher than Hindi. This highlights the higher morphological richness and orthographic variability of Bengali Romanization, motivating the use of a large shared subword vocabulary in our model.

\section{LLM Architecture}
\label{sec:model}
Our transliteration model is built upon a compact encoder–decoder transformer~\cite{vaswani2017attention, amin2024trajectoformer} model designed to handle both Hindi and Bengali Romanized inputs using a single shared architecture. We adopt the MarianNMT architecture~\cite{junczys2018marian} due to its efficient implementation, compatibility with multilingual tasks, and support for parameter sharing across source and target vocabularies. This architecture is particularly well-suited for character-level transliteration, as it captures phonetic consistency across scripts while remaining lightweight enough to train on moderately sized corpora. By leveraging shared subword vocabulary~\cite{kudo2018sentencepiece}, language-specific prefix tokens, and a unified training setup, we enable accurate and script-aware generation without needing separate models for each language.

\subsection{Model Architecture and Pretraining Strategies}
\label{subsec:training}
We frame transliteration as a conditional sequence generation problem, where the input is a Romanized character sequence $\mathbf{x} = (x_1, x_2, \dots, x_T)$ from Hindi or Bengali, and the output is the corresponding native-script character sequence $\mathbf{y} = (y_1, y_2, \dots, y_L)$. The system is trained to maximize the log-likelihood of the correct output sequence given the input, using the standard sequence-to-sequence objective:

\begin{equation}
\mathcal{L}(\theta) = -\sum_{t=1}^{L} \log p_\theta(y_t \mid y_{<t}, \mathbf{x})
\end{equation}

Here, $p_\theta(y_t \mid y_{<t}, \mathbf{x})$ denotes the probability of the $t$-th target token, computed via a softmax over the decoder hidden state: $p_\theta(y_t) = \mathrm{Softmax}(W_o h_t + b_o)$, where $W_o$ and $b_o$ are learnable output projection parameters. Training is performed using cross-entropy loss~\cite{mao2023cross} without label smoothing, preserving exact character-level supervision necessary for accurate transliteration.

We use a shared subword vocabulary vocabulary of size 32,000, trained via \textit{SentencePiece} over both Hindi and Bengali Romanized corpora This not only reduces memory overhead through tied embeddings but also improves phonetic alignment between overlapping units. A language-specific prefix token is used to distinguish tasks without altering the model structure. 

The complete model structure, including embedding dimensions, transformer depths, and feed-forward layers, is summarized in Table~\ref{tab:marian-structure}. The parameter count is constrained to approximately 60M, balancing expressiveness and generalization while remaining efficient for 27M-data training corpus.

\begin{table}[ht]
\centering
\caption{Configuration of the Marian Transformer.}
\label{tab:marian-structure}
\renewcommand{\arraystretch}{1.05}
\begin{tabular}{lcc}
\hline
\textbf{Component} & \textbf{Input} & \textbf{Output} \\
\hline
Token embedding (shared) & 32k vocabulary $\times$ 512 & 512 \\
Positional encoding       & 512 & 512 \\
\hline
Encoder $\times$6         & 512 & 512 \\
\quad Self-attention (8-head) & 512 & 512 \\
\quad Feed-forward         & 512 & 2048 $\rightarrow$ 512 \\
\hline
Decoder $\times$6         & 512 & 512 \\
\quad Self-attention (8-head) & 512 & 512 \\
\quad Encoder–decoder attention & 512 & 512 \\
\quad Feed-forward         & 512 & 2048 $\rightarrow$ 512 \\
\hline
Output projection (LM head) & 512 & 32k vocabulary \\
\hline
\end{tabular}
\end{table}

For regularization, we disable label smoothing to preserve one-to-one character mappings. Instead, we apply $\ell_2$ weight decay (0.01), dropout (0.1), and global gradient clipping (threshold 1.0). A low learning rate ($2 \times 10^{-5}$) with linear warm-up over the first 1,000 steps followed by cosine decay ensures stable convergence in low-resource settings. This configuration yields a compact, multilingual transliteration system capable of handling Romanized Hindi and Bengali inputs using a single model with shared parameters, strong generalization, and efficient training.

\subsection{Implementation and Experimental Setup}
\label{subsec:implementation}

Experimental codes are implemented using \textit{PyTorch} and \textit{HuggingFace Transformers}. Training is performed on a workstation with an NVIDIA RTX 4090 (16 GB), 32 GB RAM, and a 13th Gen Intel Core i9 CPU.

The tokenized sequences are padded or truncated to 128 tokens. We use the \texttt{Seq2SeqTrainer} API for end-to-end management of training and evaluation. The model and tokenizer share the same SentencePiece vocabulary, enabling uniform handling of Hindi and Bengali inputs. Hyperparameter settings are detailed in Table~\ref{tab:training-settings}, tuned to balance computational efficiency with learning stability.

\begin{table}[ht]
\centering
\caption{Training configuration and hyperparameters.}
\label{tab:training-settings}
\renewcommand{\arraystretch}{1.05}
\begin{tabular}{lc|lc}
\hline
\textbf{Parameter} & \textbf{Value} & \textbf{Parameter} & \textbf{Value} \\
\hline
Batch size & 128 & Attention heads & 8 \\
Max seqe length & 512 & Hidden size & 512 \\
Encoder / Decoder & 6 / 6 & FFN dimension & 2048 \\
Learning rate & $2 \times 10^{-5}$ & Weight decay & 0.01 \\
Epochs & 5 & Optimizer & AdamW \\
Dropout & 0.1 & Gradient clipping & 1.0 \\
Tokenizer & SentencePiece & Vocabulary size & 32,000 \\
\hline
\multicolumn{3}{l}{\textbf{Total Parameters:}} & \textbf{60M} \\
\hline
\end{tabular}
\end{table}

Evaluation is conducted using corpus-level BLEU and character error rate (CER), capturing both sequence fluency and character-level fidelity.

\section{Results and Analysis}
\label{sec:results}
In this section, we present and analyze the performance of our Marian architecture inspired Transliteration LLM on the IndoTranslit dataset, which includes transliteration pair for Hindi, Bengali, and a combined multilingual setting (Multi-Trans). Our evaluation highlights both the quantitative and qualitative strengths of the model, demonstrating its ability to accurately convert Romanized input into native scripts in a multilingual context.

 Table~\ref{tab:bengali} shows the top three most accurate and least accurate predictions. The accurate examples demonstrate the model's ability to produce fluent and fully correct native script translations from Romanized inputs. On the other hand, the least accurate predictions show cases where the model struggles due to ambiguity or rare phrasing. These include structurally varied or partially code-mixed inputs such as \textit{'aj ke class nai'} or \textit{'kya hal hai'}, where the model often outputs near-matching but semantically imperfect predictions. These examples illustrate common challenges in real-world transliteration scenarios and offer insight into potential areas for future improvement.

\begin{table*}[ht]
    \centering
    \caption{Most accurate and inaccurate predictions of the model on the test set.}
    \label{tab:bengali}
    \begin{tabular}{c}
        \includegraphics[width=0.95\linewidth]{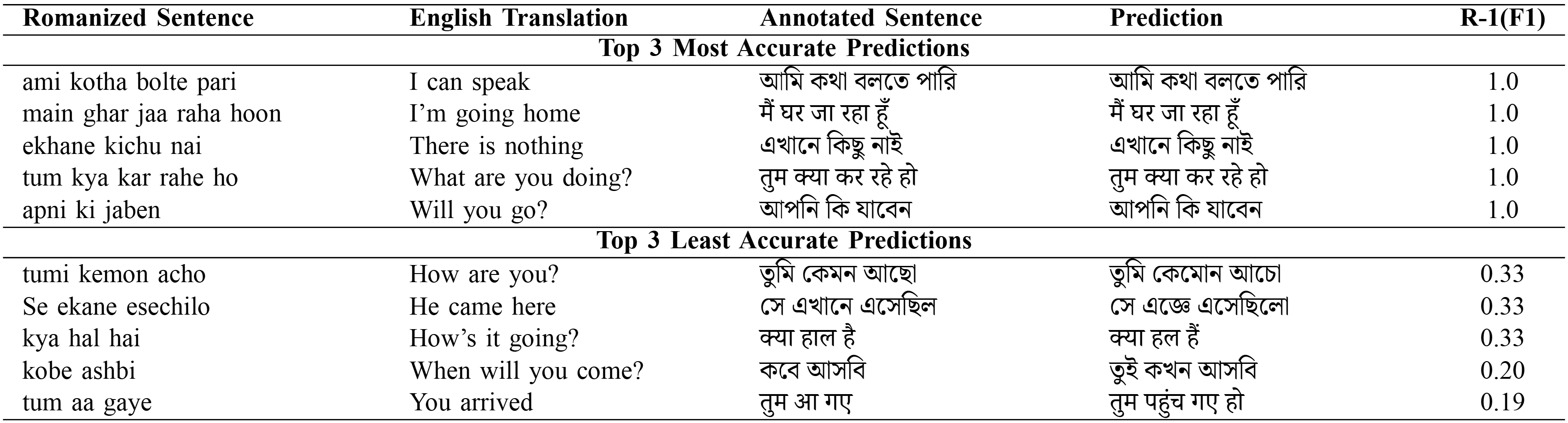}
    \end{tabular}
\end{table*}

\begin{table*}[!ht]
\footnotesize
\centering
\caption{Model benchmarking on \textbf{Hindi} and \textbf{Bengali} transliteration tasks.}
\begin{tabular}{llcccccccc}
\hline
\textbf{Dataset} & \textbf{Model \&} & \multicolumn{3}{c}{\textbf{ROUGE Score}} & \textbf{BLEU} & \textbf{CER} & \textbf{WER} & \textbf{chrF} & \textbf{METEOR} \\
\cmidrule(lr){3-5}
                 &   \textbf{Op-Mode} & \textbf{R-1} & \textbf{R-2} & \textbf{R-L} & \textbf{(Corpus)} & \textbf{Score} & \textbf{Score} & \textbf{Score} & \textbf{Score} \\
\hline
\multirow{4}{*}{IndoTranslit} & \textbf{Ours (Marian)} & & & & & & & & \\
& Hindi           & 54.41 & 41.78 & 54.44 & 77.57 & 0.09 & \textbf{0.13} & \textbf{88.39} & 91.61 \\
& Bengali          & 48.59 & \textbf{47.19} & 48.57 & \textbf{77.82} & 0.16 & 0.25 & 86.13 & 80.12 \\
& \textbf{Multi-Trans} & \textbf{58.59} & 43.46 & \textbf{58.58} & 73.15 & \textbf{0.08} & 0.14 & 88.01 & \textbf{92.43} \\
\hline
\end{tabular}
\label{tab:ours_translation_scores}
\end{table*}

Table~\ref{tab:ours_translation_scores} presents detailed performance metrics across different evaluation settings using standard scores such as ROUGE, BLEU, CER, WER, chrF, and METEOR~\cite{ahmadnia2018neural, khan2021end, fahim2024banglatlit}. Our model shows strong performance across all categories. For Hindi, we achieve a BLEU score of 77.57 with low CER (0.09) and WER (0.13), indicating highly accurate character and word-level predictions. Bengali performance is similarly strong, with an even higher BLEU score of 77.82 but slightly higher error rates, reflecting the more complex spelling variations commonly found in Romanized Bengali. Importantly, the Multi-Trans configuration, where a single model handles both languages, achieves balanced performance with a BLEU score of 73.15 and competitive scores across all other metrics. These results confirm that a multilingual Seq2Seq model works well without needing separate models for each language.

\begin{figure}[ht]
    \centering
    \includegraphics[width=\linewidth]{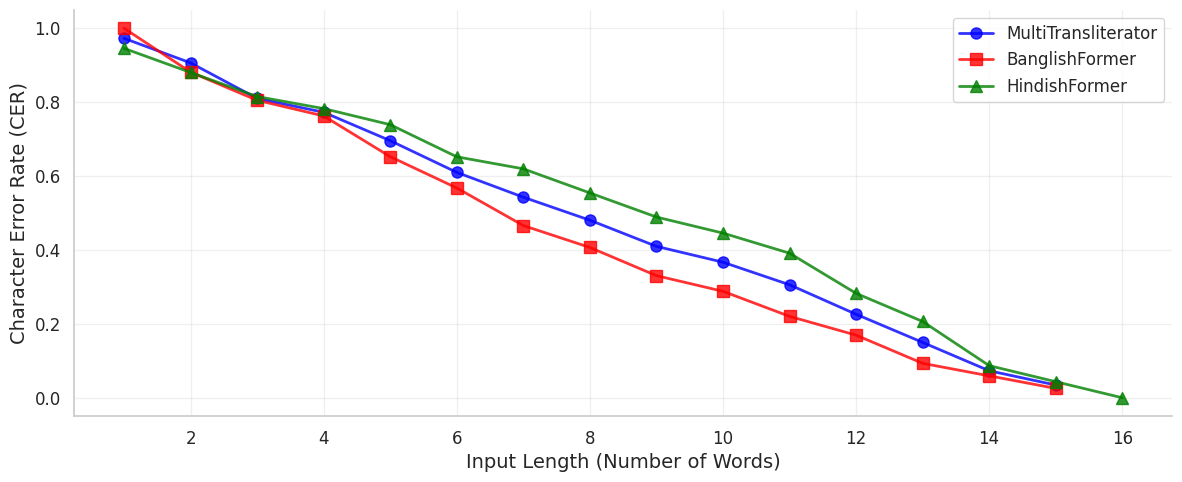}
    \caption{Sentence length vs. average CER score for Hindi, Bengali, and Multi-Trans models}
    \label{fig:cer_length_comparison}
\end{figure}

Figure~\ref{fig:cer_length_comparison} shows how the average Character Error Rate (CER)~\cite{park2024character} varies with input length for three models. All systems improve as sentence length increases, likely due to greater contextual information. The MultiTransliterator maintains strong performance throughout, closely matching the single-language BanglishFormer and HindishFormer models. Notably, shorter inputs (under 5 words) lead to higher CER across all models, reflecting ambiguity in short transliteration sequences.

\begin{table}[!ht]
\footnotesize
\centering
\caption{Comparison of transliteration benchmarks.}
\begin{tabular}{lllll}
\hline
\textbf{Paper} & \textbf{Testset} & \textbf{Score} & \textbf{Model} & \textbf{Metric} \\
\hline
Aksharantar & Dakshina & 60.6 & IndicXlit & Char-BLEU \\
BanglaTLit  & BanglaTLit & 78.9 & TB-XLM-R+B5 & Char-BLEU \\
IndoNLP     & IndoNLP & 62.8 & LLaMA 3.1 & Word-BLEU \\
IndiMiner   & Dakshina & 62.8 & LLaMA+LoRA & BLEU-W \\
BanTH-Bin   & BanTH & 82.6 & TB-mBERT & Accuracy \\
BanTH-Mul   & BanTH & 30.2 & TB-BERT & Macro-F1 \\
\textbf{Ours (Mul)}        & \textbf{IndoTranslit} & \textbf{73.15} & \textbf{Marian} & \textbf{BLEU (C)} \\
\hline
\end{tabular}
\label{tab:result_comparison}
\end{table}

We also compare our results with relevent existing benchmarks in Table~\ref{tab:result_comparison}. Our model achieves a BLEU score~\cite{tran2019does} of 73.15 on the IndoTranslit test set, outperforming IndicXlit (60.6 on Dakshina), LLaMA-based models (62.8), and BanglaTLit (78.9). Even when compared with binary classification (BanTH-Bin: 82.6 Accuracy) and macro-F1 setups (BanTH-Mul: 30.2), our score remains significantly higher, specially given that our model handles generation at the multilingual prompt. These comparisons show that IndoTranslit, combined with our Marian-based multilingual model, sets a new strong benchmark for Romanized-to-native transliteration in Hindi and Bengali. Future work will explore compact student–teacher setups and multi-stage distillation to extend IndoTranslit toward low-resource Indic languages, building on advances in efficient model replication~\cite{gharami2025clone}.

\section{Conclusion}
\label{sec:conclusion}
This paper presented IndoTranslit, a large-scale dataset and multilingual Seq2Seq model tailored for transliteration of Romanized Hindi and Bengali text. The dataset comprising over 2.7 million aligned pairs, captures both clean and noisy variations, effectively bridging the gap between formal phonetic transliteration and informal user-generated writing. Our Marian-based multilingual LLM demonstrates robust generalization across Hindi and Bengali, outperforming existing models such as IndicXlit and LLaMA-based baselines on BLEU, CER, and WER metrics while maintaining a lightweight, deployment-friendly configuration. The shared subword vocabulary and unified architecture prove effective for cross-lingual phonetic alignment and consistent transliteration across scripts.

Overall, IndoTranslit advances transliteration research by addressing three key gaps: (1) the lack of large-scale, mixed-quality Romanized datasets; (2) the absence of multilingual training for Indo-Aryan scripts; and (3) the limited adaptability of current models to noisy, real-world text. Future work will extend this effort to additional Indo-Aryan languages, explore context-sensitive and code-mixed modeling, and develop compact on-device transliteration modules to support real-time multilingual applications across South Asian digital ecosystems.

% This paper introduces IndoTranslit, a large-scale transliteration dataset, and a multilingual Seq2Seq language model designed to handle Romanized Hindi and Bengali scripts. Our 2.7 million transliteration pairs incorporate both clean and informal variations, enabling realistic and diverse model training. The developed Marian-based model demonstrates strong performance across multiple evaluation metrics, outperforming existing baselines while remaining lightweight and deployment-friendly. Future work will focus on extending IndoTranslit to include a broader range of prominent Indo-Aryan languages to enhance the dataset’s linguistic diversity and improve its effectiveness for multilingual large language model applications across South Asia.

\bibliographystyle{IEEEtran}
\bibliography{references1}

\end{document}